%% file: main.tex
\documentclass[a4paper, oneside, twocolumn, notitlepage, 10pt]{extarticle_ecoc}
\usepackage{ecoc}
\usepackage{acronym}
\usepackage{multirow}
\usepackage{enumerate}
\input{acronyms.tex}
\usepackage{soul}
\begin{document}

\selectlanguage{english}    


\title{Multi-Span Optical Power Spectrum Evolution Modeling using ML-based Multi-Decoder Attention Framework}
\vspace{-8mm}

\author{
    Agastya Raj\textsuperscript{*,}\textsuperscript{(1)}, 
    Zehao Wang\textsuperscript{(2)},
    Frank Slyne\textsuperscript{(1)},
    Tingjun Chen\textsuperscript{(2)}, 
    Dan Kilper\textsuperscript{(1)}, 
    Marco Ruffini\textsuperscript{(1)}
}
\vspace{-6mm}
\maketitle 
\begin{strip}
    \begin{author_descr}

        \textsuperscript{(1)} CONNECT Centre, School of Computer Science and Statistics and School of Engineering, Trinity
College Dublin, Ireland
        \textcolor{blue}{\uline{*rajag@tcd.ie}}

        \textsuperscript{(2)} Duke University, Department of Electrical and Computer Engineering, Durham, NC, USA

    \end{author_descr}
    \vspace{-4.3mm}
\end{strip}

\renewcommand\footnotemark{}
\renewcommand\footnoterule{}

\begin{strip}
    \begin{ecoc_abstract}
        We implement a ML-based attention framework with component-specific decoders, improving optical power spectrum prediction in multi-span networks. By reducing the need for in-depth training on each component, the framework can be scaled to multi-span topologies with minimal data collection, making it suitable for brown-field scenarios. 
        \textcopyright2024 The Author(s)
    \end{ecoc_abstract}
    \vspace{-4.7mm}
\end{strip}

\section{Introduction}
\vspace{-1.5mm}
Today’s network operators depend on advanced networks, which include \ac{ROADM} systems using flex-grid \ac{DWDM}, for high-speed and low latency services. As these networks become more configurable and adaptable, accurate estimation of optical link performance, including power spectrum evolution and \ac{OSNR}, has become crucial~\cite{lowmargin}. \ac{DWDM} signals experience different propagation characteristics through various components, for example due to the wavelength dependent gain of \acp{EDFA} and losses induced by fibers and \acp{WSS}. The challenge largely arises from the difficulty in developing accurate models of devices like \acp{EDFA}, where different units can show significant variations, for example, in their wavelength-dependent gain. 

Traditional monolithic end-to-end system models, when applied to a large network, require extensive in-field data collection and may overlook the nuanced interactions between components, which are important for understanding power evolution through the network~\cite{Kruse_lstm,Morette_dt}. In addition, these models are limited in the sense that any network topology change will require training of new models, with extensive collection of new data\cite{liu2019hierarchical}. It has been shown that individual components such as \acp{EDFA} can be characterized using \ac{ML} to predict optical power spectrum. However, in multi-span networks involving multiple such components, a direct cascade of these models perform poorly, resulting in high error accumulation~\cite{9979538,9605932}. Recently, a \ac{CL} framework was proposed using component-level models for each \ac{EDFA} in a multi-span network, utilizing additional end-to-end measurements~\cite{wang_multi_span}. This method enabled models trained in the lab to be applied in the field as a scalable approach for large networks. However, this process requires detailed characterization of each \ac{EDFA} before deployment in order to achieve low end-to-end prediction errors. This motivates the challenge of finding the right balance of lab data collection and field measurements--minimizing the quantity and complexity of the field measurements while achieving high accuracy. 




In this paper, we introduce a novel approach by interpreting a network as a series of data points from \acp{OCM} deployed within in-line \acp{ROADM}. We propose a sequential Multi-Decoder Attention Model (MDAM) that leverages intermediate data to accurately emulate power spectrum evolution across a network. This is achieved by encoding the input signal with a shared attention-based \ac{LSTM} encoder, and utilizing component-specific decoders to predict the power spectrum at each network node, improving power spectrum evolution prediction in multi-span networks. Instead of cascading discrete component-models, the encoder maintains a shared information layer through the model, while decoders enable accurate predictions for specialized component modeling in a unified and scalable framework.
\footnote{This paper is a preprint of a paper submitted to ECOC 2024 and is subject to Institution of Engineering and Technology Copyright. If accepted, the copy of record will be available at IET Digital Library}

Another key issue in modeling power spectrum evolution is the limited number of measurements that might be available for green field and brown field scenarios~\cite{lowmargin}. In brown field scenarios, working with fewer measurements is the key because of the high cost/complexity associated with live network probing, as these could introduce impairments on existing live channels. In green field scenarios, using less training data speeds up the process of accurately modeling each device. To enable high adaptability and ensure minimal data collection in these scenarios, we introduce a novel \ac{TL} process. We first develop a base model of encoder and device-specific decoders within a controlled laboratory setup, featuring a single physical device for each component type. 
\ac{TL} is then applied to the other devices in a multi-span topology to scale up the model. 
This strategy can, for example, be used to first carry out in-depth data collection in a laboratory environment, and reduce the number of data points from the live network. This greatly simplifies model training and deployment, requiring in-depth characterization of only single physical devices to predict power spectra for unseen, multi-span networks.


In order to study the performance of MDAM and its ability to carry out transfer learning across different networks, we carry out experiments across two different testbeds. A first model pre-training in OpenIreland~\cite{openireland} is followed by a transfer towards two multi-span target topologies, over 234 km of fiber, in the COSMOS PAWR Testbed~\cite{cosmos}. Experimental results show that our approach achieves a 50-fold reduction in the amount of training data with respect to a state of the art benchmark model ~\cite{wang_multi_span}, while also improving prediction accuracy. 


\begin{figure*}[t]
	\centering
    \includegraphics[width=\linewidth]{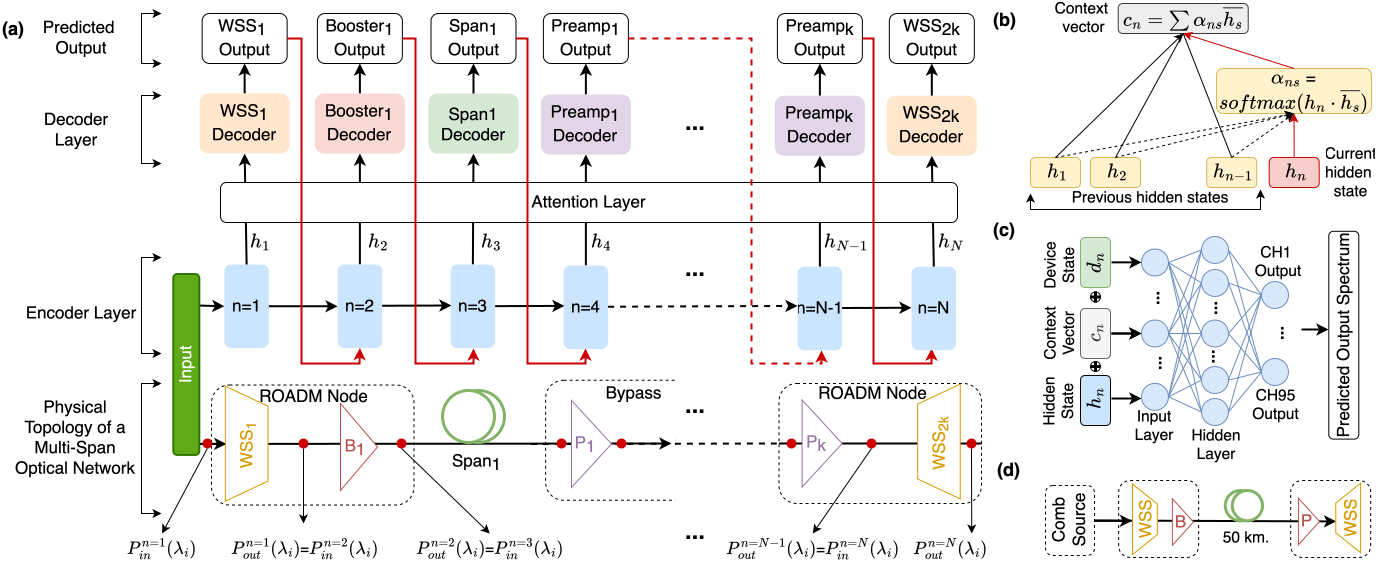}
	\caption{(a) Multiple-Decoder Attention Model (MDAM) architecture, (b) Dot-product attention layer mechanism for \(n\)\textsuperscript{th} component, (c) Component-specific decoder model structure, (d) Base model training setup}
	\label{model}
 \vspace{2mm}
\end{figure*}

\begin{table*}[t]
\tiny \tiny \tiny
    \centering
    \renewcommand{\arraystretch}{1.2}
    \setlength{\tabcolsep}{3pt}
    \resizebox{\textwidth}{!}{
    \begin{tabular}{c|ccc|ccc} \hline 
    
         \multirow{2}{*}{\textbf{End-component absolute prediction error: Mean/95\textsuperscript{th} percentile (dB)}}&  \multicolumn{3}{c|}{\textbf{Random}} & \multicolumn{3}{c}{\textbf{Goalpost}}\\ \cline{2-7}

         & \textbf{Direct Cascade} & \textbf{Benchmark} & \textbf{MDAM} & \textbf{Direct Cascade}&  \textbf{Benchmark} & \textbf{MDAM}\\ \hline 
         
         Topo. \#1 (6-span, 234 km): 40 km–40 km–40 km–\textit{32 km}–\textit{32 km}–50 km& 1.73/4.39 & 0.16/0.43&  0.16/0.27& 2.13/2.60 & 0.59/1.48& 0.20/0.49\\ \hline 
         
         Topo. \#2 (4-span, 234 km): 40 km–\textit{72 km}–\textit{72 km}–50 km&  0.61/2.03 & 0.16/0.42& 0.14/0.21& 1.03/1.28 & 0.58/1.43& 0.24/0.47\\ \hline
         
    \end{tabular}}
    \caption{Mean/95\textsuperscript{th} percentile absolute error of end-component predictions for different topologies (\textit{italics} spans indicate field fibers)}
    \label{table}
    \vspace{-6mm}
\end{table*}

\vspace{-3.5mm}
\section{Model Architecture}
\vspace{-0.3mm}

\begin{figure*}[t]
    \centering
    \includegraphics[width=\linewidth]{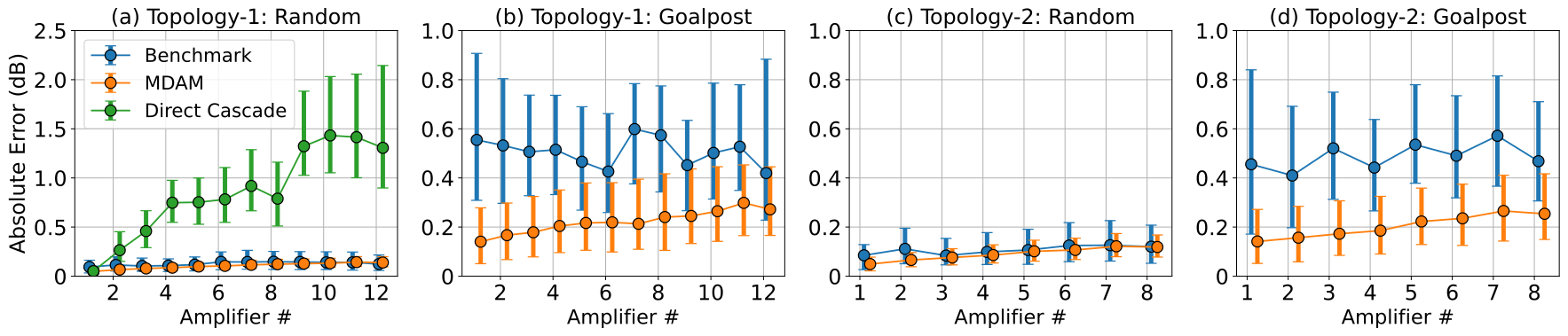}
    \caption{Comparison of absolute error distribution in power evolution predictions for Benchmark vs MDAM model for Topology-1 in (a) Random, (b) Goalpost, and for Topology-2 in (c) Random and (c) Goalpost configurations. Markers denote the median, and the whiskers denote the inter-quartile range (25\textsuperscript{th}-75\textsuperscript{th} percentile). \textit{(Comparison with Direct Cascade model shown only in (a) for clarity)}}
    \label{results}
    \vspace{-2mm}

\end{figure*}



The data from \acp{OCM} in a transmission system with \(N\) components comprises a sequence of power spectrum values for each component~\cite{wang_edfa_data} \(\overline{P}{(\lambda_i)} = [P_0{(\lambda_i)}, P_1{(\lambda_i)}, P_2(\lambda_i), ... P_N{(\lambda_i)}]\), which defines the environment. \(P_0{(\lambda_i)}\) is the input power spectrum at first \ac{ROADM}'s \ac{EDFA}, and \(P_n{(\lambda_i)}, n \in \{0, 1, \cdots, N\}\) denotes the power spectrum after component \(n\). Given the initial power spectrum \(P_0\), our objective is to predict power spectra after transmission through each component in the network. Since the output of each component serves as an input to the next component, this can be treated as an auto-regressive sequential modeling problem. For any \(n^{th}\) component, the input features \(P_{in}{(\lambda_i)}\) and output features \(P_{out}{(\lambda_i)}\) can be defined as \(P_{n-1}{(\lambda_i)}\) and \(P_{n}{(\lambda_i)}\) respectively. 


We employ a sequential model architecture using a shared encoder and multi-decoder model with attention layer (refer to Fig.~\ref{model}(a)). The encoder is a 3 layer \ac{LSTM} model with a hidden size of 100 units each, with a dropout of 20\% applied at each layer to reduce overfitting. Given the substantial variations in the physical behavior of different network devices, a single shared output layer proves inadequate for accurately modeling each component. Instead, we create component-specific decoders for each component used, namely, Booster, Preamp, Span and \ac{WSS}. As shown in Fig.~\ref{model}(c), decoders are shallow neural networks consisting of a single hidden layer of 100 neurons, and 95 neurons in the final layer, predicting the channel-wise power output. A deeper encoder helps the model accommodate diverse components across longer networks, while shallow decoders are sufficient to model a single component~\cite{kong_multilingual_2021}. Dot-product attention is implemented to avoid error accumulation and dynamically focus on signal's evolution through the network, without adding parameter overhead~\cite{vaswani2017attention}.
As the signal progresses, at any step \(n\), the encoder processes the output from previous step, generating a hidden state vector \(h_n\). The attention layer (refer Fig.~\ref{model}(b)) computes attention scores for all previous components using dot product attention where \(score(n) = h_n \cdot \overline{h_s}\), with \(\overline{h_s} = [h_1, h_2, \cdots h_{n-1}]\) denoting all the previous hidden states. Softmax is applied to these scores to derive attention weights \(\mathbf{{\alpha_{ns}}}\), and a context vector \({c_n}\) is calculated as a weighted sum of all previous hidden states weighted by their respective attention weights, give by \({c_n}=\sum {\mathbf{\alpha_{ns}}}\overline{h_s}\). This context vector is concatenated along with the current hidden state and corresponding device configuration features \(\mathbin{d_n}\)(such as \ac{EDFA} target gain, span length and \ac{WSS} attenuation), to get the final output vector \(o_n = h_n\mathbin\oplus{c_n}\oplus{d_n}\). 
This output vector is then passed to the corresponding decoder to get the predicted power spectrum for that component. This predicted power spectrum serves as the input for next component step.

We develop the shared encoder and base models for four decoders types, each corresponding to a different component type (Booster, Preamp, Span, and WSS). This is implemented in a laboratory environment in the OpenIreland Testbed~\cite{openireland}, shown in Fig.~\ref{model}(d), where we carry out a total of 3,168 power spectrum measurements. 
The model undergoes a two-phase training process: \textbf{a. Teacher-Forcing Phase}: for the initial 3,000 epochs, training uses ground-truth power spectra to accelerate learning and reduce error propagation~\cite{rnn_tf}. \textbf{b. Auto-Regression Phase}: the model is trained in auto-regressive mode for 9,000 epochs. Throughout both processes, \ac{SGD} optimizer is used to optimize the total weighted \ac{MAE} across all components~\cite{keskar2017improving}, with an initial rate of 1e-03, decaying exponentially every 1,000 epochs. The encoder and decoders are preserved separately for subsequent transfer learning. 

\vspace{-2.7mm}
\section{Experimental Setup and Transfer Learning}

To evaluate the performance of MDAM, we employ the PAWR COSMOS testbed in Manhattan, USA~\cite{cosmos} as the target network, conducting transfer learning with base models developed in OpenIreland and then collect the performance results.
We set up two multi-span topologies: one 6-span link with 12 \acp{EDFA} and one 4-span link with 8 \acp{EDFA}. The span configuration for the two topologies is summarized in Table~\ref{table}, with total lengths of 234 km (including 64 km of Manhattan field fibers). For both topologies, all booster and pre-amplifier \acp{EDFA} are set to high gain mode with 18 dB gain and zero gain tilt. A comb source is used to generate 95$\times$50 GHz \ac{WDM} channels in the C-band, and channel configurations and spectrum flattening are managed at the initial MUX, with the signal traversing all spans and being dropped at the end DEMUX. We consider three types of channel loading configurations with different number of loaded channels: \textit{(i)} Fixed (fully/half loaded), \textit{(ii)} Random (loading with randomly selected channels), and \textit{(iii)} Goalpost (groups of channel loading in different spectral bands).
For transfer learning, we employ the pre-trained base encoder and replicate component specific decoders--assigning them to corresponding devices in the target network. In total, we use 48 measurements (fixed and random loading) for transfer learning into the COSMOS target network, using the same two-phase training process described in the previous section. However, we use a reduced initial learning rate of 1e-5 and a lower gradient clipping threshold of 0.5. 




\vspace{-2.7mm}
\section{Results} 
\vspace{-0.7mm}

We compare our model against a Direct Cascade of individually trained component-level models for each \ac{EDFA} in the network, and the benchmark CL-Model\cite{wang_multi_span}; with the test set consisting of 658 random and 27 goalpost data points. Table~\ref{table} summarizes the mean and 95\textsuperscript{th} percentile absolute errors of end-\ac{EDFA} power spectrum prediction for two topologies across random and goalpost configurations. It can be seen that Direct Cascade suffers high accumulation of errors, while CL-Model and MDAM achieves a similar MAE for random configuration. However, MDAM displays improved predictions in extreme/edge cases, showing a lower 95\textsuperscript{th} percentile error. Moreover, MDAM outperforms the benchmark model in goalpost configuration, showing a more stable distribution of errors across diverse channel configurations. This improvement is particularly significant, given the reduction in measurements from over 160,000 in the direct cascade model and the benchmark study~\cite{wang_multi_span} to 3,216 in this work (of which only 48 is in the target network).

Fig.~\ref{results} shows the distribution of absolute prediction errors across both topologies for random and goalpost configurations. MDAM displays a reduced error accumulation through the network and a more stable error distribution. Especially in the goalpost scenario, MDAM outperforms the benchmark model with a consistent median absolute error\(<\)0.3 dB through all components. Note that MDAM jointly predicts power spectrum for all components along the network, other methods require separate models to be trained for individual devices. Additionally, MDAM can be fine-tuned with added measurements for network expansions, while other models require complete retraining for new network configurations.

\vspace{-3.5mm}
\section{Conclusion}
\vspace{-0.5mm}
We demonstrate a scalable ML-based model pre-trained on devices of the same manufacturer that can be generalized and transferred to a larger network. Our results show improved performance with respect to a state-of-the-art benchmark model, while achieving a 50-fold reduction in training data. 

\clearpage
\section{Acknowledgements}
The work was supported by SFI grants 12/RC/2276-p2, 22/FFP-A/10598, 18/RI/5721 and  13/RC/2077-p2, and NSF grants CNS-1827923, OAC-2029295, CNS-2112562, CNS-2211944, and CNS-2330333.
\printbibliography

\end{document}

%% file: acronyms.tex
\acrodef{EDFA}{Erbium-Doped Fiber Amplifier}
\acrodef{DWDM}{Dense Wavelength Division Multiplexing}
\acrodef{ROADM}{reconfigurable optical Add-Drop Multiplexer}
\acrodef{MAE}{Mean Absolute Error}
\acrodef{ML}{Machine Learning}
\acrodef{WSS}{Wavelength Selective Switch}
\acrodefplural{WSS}{Wavelength Selective Switches}
\acrodef{WDM}{Wavelength Dependent Multiplexing}
\acrodef{OCM}{Optical Channel Monitor}
\acrodef{LSTM}{Long Short-Term Memory}
\acrodef{OSNR}{Optical Signal-to-Noise Ratio}
\acrodef{CL}{Cascaded Learning}
\acrodef{MDAM}{Multi-Decoder Attention Model}
\acrodef{SGD}{Stochastic Gradient Descent}
\acrodef{TL}{Transfer Learning}